\documentclass[11pt]{article}

\usepackage[letterpaper, margin=1in]{geometry}
\usepackage[utf8]{inputenc}
\usepackage[T1]{fontenc}
\usepackage{times}

\usepackage[numbers,compress]{natbib}

\usepackage{hyperref}
\usepackage{url}

\usepackage{booktabs}
\usepackage{multirow}
\usepackage{float}
\usepackage{enumitem}
\usepackage{caption}
\usepackage{subcaption}
\usepackage{amsmath}
\usepackage{amsfonts}
\usepackage{amssymb}
\usepackage{nicefrac}

\usepackage{microtype}
\usepackage{xcolor}
\usepackage{graphicx}

\title{\vspace{-2.0em}\Large\bfseries ENVS: Environment-Native Verified Search for Long-Horizon GUI Agents
}

\author{
Yincheng Zhou$^{1,*}$ \quad Athena Zhuoming Zhong$^{1,*}$ \quad Shijie Zhang$^{2,*}$\\
Kevin Zhang$^{2}$ \quad Teresa Xiaotao Shang$^{1}$ \quad Shanghang Zhang$^{2,\dagger}$\\[0.5em]
\textit{$^{1}$University of Pennsylvania \quad $^{2}$Peking University}\\[0.2em]
\small $^{*}$Equal contribution. \quad $^{\dagger}$Corresponding author.
}
\date{}
\begin{document}
\maketitle
\begingroup\renewcommand\thefootnote{}\footnotetext{Code: \url{https://github.com/ArtysicistZ/ENVS}}\endgroup
\vspace{-1.85em}     

\begin{figure}[H]
\centering
\includegraphics[width=1\linewidth]{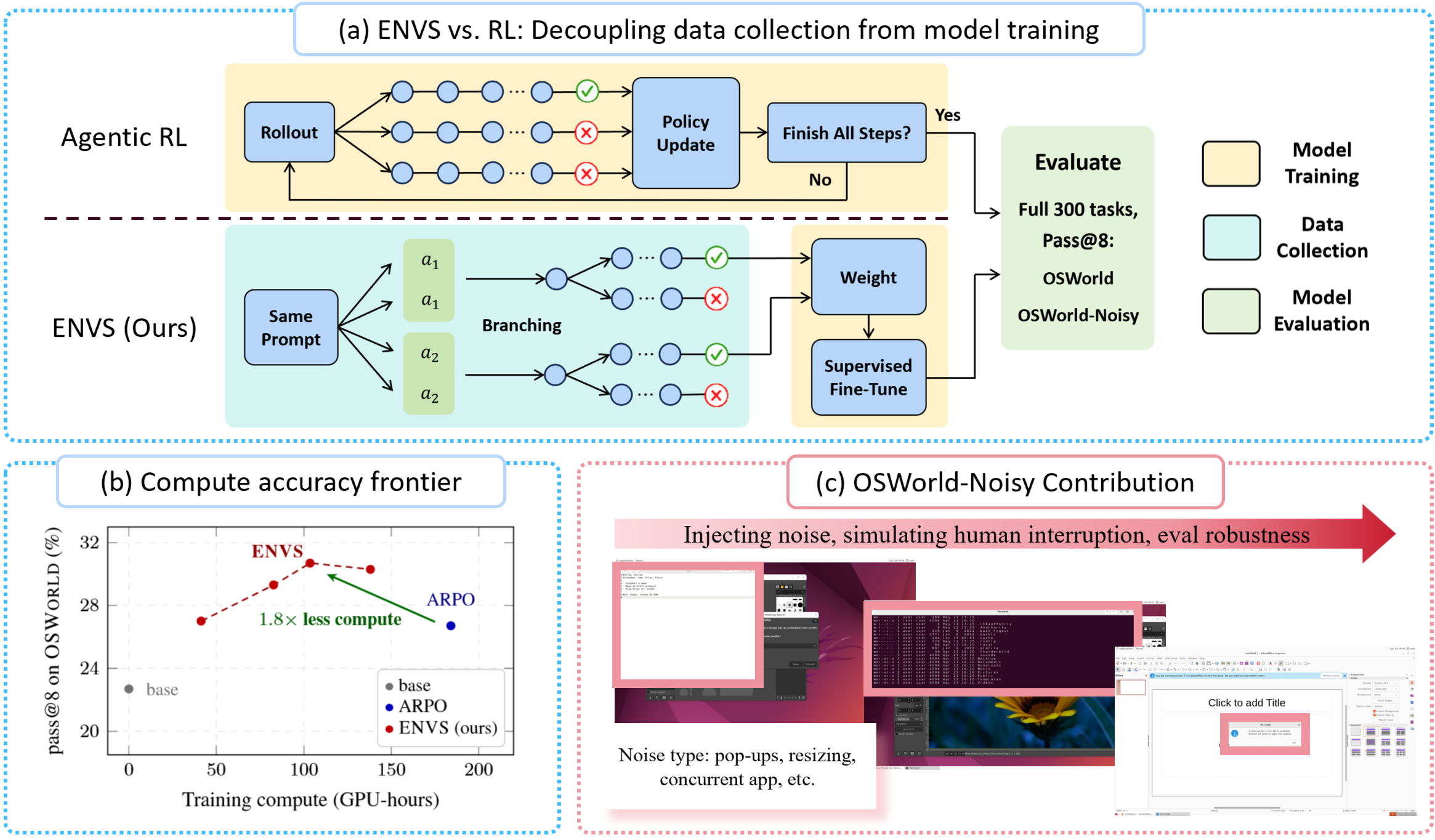}
\caption{\textsc{ENVS} decouples data collection from model training \textbf{(a)}, reaching higher accuracy at lower compute than online RL \textbf{(b)}; \textsc{OSWorld-Noisy} injects human-style interruptions to test robustness \textbf{(c)}.}
\label{fig:teaser}
\end{figure}

\begin{abstract}
As multimodal agents move from interface understanding to real software control, successful trajectory discovery in live desktop environments becomes a key challenge. GUI tasks require long-horizon sequences of precise mouse and keyboard actions, while feedback is sparse, delayed, and costly to obtain through VM rollouts. We propose Environment-Native Verified Search (\textsc{ENVS}), a training-time search-and-filter pipeline that uses the environment to construct verified supervision before policy optimization: it branches over behaviorally distinct GUI actions in live \textsc{OSWorld} VMs, verifies successful leaves, and trains from globally balanced step-level supervision. To evaluate robustness under realistic desktop interruptions, we also introduce \textsc{OSWorld-Noisy}, a dynamic benchmark for recoverable desktop interruptions that preserves the original tasks while testing whether agents can refocus, dismiss, wait, or recover under live perturbations. On the 300-task \textsc{OSWorld} pool, \textsc{ENVS} reaches 30.3 pass@8 on original evaluations and 29.0 on \textsc{OSWorld-Noisy}, outperforming matched ARPO-style online RL while reducing compute from 184--192 to 138--153 GPU-hours; even with only 30\% of its search data, \textsc{ENVS} reaches 27.0 pass@8, exceeding ARPO from the base model. Training from noisy environments also better preserves visual-reasoning abilities on auxiliary benchmarks, including \textsc{OSWorld-G Refusal} (16.7 vs. 1.9) and \textsc{BLINK Functional Correspondence} (26.2 vs. 23.1).

\end{abstract}

\section{Introduction}
Multimodal foundation models are increasingly being used as agents that operate real software through screenshots, mouse actions, and keyboard actions~\citep{cheng2024seeclick,hong2023cogagent,zhang2024screenagent,qin2025uitars}. GUI agents are a demanding testbed for this progress because they require visual grounding, long-horizon planning, recovery from mistakes, and reliable execution in stateful environments.

Training such agents is difficult because success feedback is sparse, delayed, and expensive to obtain. In \textsc{OSWorld}, each rollout executes inside a live desktop VM, and the environment usually provides only terminal task success or failure~\citep{xie2024osworld}. A failed trajectory often does not reveal which earlier action caused the failure. Thus, a central bottleneck is not only how to optimize a policy, but how to discover enough high-quality successful trajectories in the first place.

Recent work addresses sparse agent feedback through online RL and search. ARPO-style methods adapt GRPO with replay or rollout modifications for GUI and tool-use agents~\citep{lu2025arpo_gui,dong2025agentic_arpo,qian2025toolrl,wang2025ragen}, while tree-structured methods such as SEEA-R1~\citep{tian2025seea} integrate MCTS-style exploration with online policy optimization for long-horizon embodied agents. Search-based language-agent methods also explore multiple candidate reasoning or action paths before committing to a solution~\citep{yao2023treeofthoughts,koh2024tree_search_lm_agents,putta2024agentq}. These methods motivate ENVS, but they still use environments mainly for online rollouts or inference-time search rather than for verified, balanced data construction before training. In online optimization, trajectory discovery remains coupled to policy updates: easier tasks can dominate the successful rollout distribution, and replay or clipping only improves local update stability without directly controlling the global allocation of gradient across tasks and trajectory lengths.

We propose \textsc{ENVS} (\emph{Environment-Native Verified Search}). Rather than introducing a new reinforcement learning algorithm, \textsc{ENVS} is a search-based training-data construction framework that leverages environment-native verification signals to identify successful trajectories. Instead, it separates trajectory discovery from policy optimization. It searches in live \textsc{OSWorld} VMs, branches over behaviorally distinct executable actions, verifies trajectory leaves with the \textsc{OSWorld} oracle, and converts successful trajectories into globally balanced step-level supervised data. Unlike classical MCTS or Tree-GRPO, \textsc{ENVS} uses search only for training-time trajectory discovery, not online policy improvement. Because training happens after collection, \textsc{ENVS} can cap overrepresented easy tasks, upweight rare-success tasks, normalize long trajectories, and control how much gradient mass each part of the discovered dataset receives. 

We also introduce \textsc{OSWorld-Noisy}, a benchmark for evaluating GUI agents under recoverable live desktop interruptions. It preserves the original \textsc{OSWorld} tasks and evaluators while injecting controlled pop-ups, focus changes, dialogs, overlays, and background activity that require agents to refocus, dismiss, wait, or recover before continuing.
We evaluate \textsc{ENVS} on the 300-task \textsc{OSWorld} evaluation pool, consisting of 86 trainable tasks and 214 held-out tasks, using the same pass@8 protocol as the ARPO-style baselines. 

Our main findings are:

\begin{itemize}[leftmargin=*, itemsep=0.15em, topsep=0.15em, parsep=0pt, partopsep=0pt]
    \item \textbf{Clean performance:} \textsc{ENVS} reaches 30.3 pass@8 on \textsc{OSWorld}, improving UI-TARS-1.5 from 22.7 and outperforming completed ARPO-style baselines at 26.7.
    \item \textbf{Noisy robustness:} On \textsc{OSWorld-Noisy}, \textsc{ENVS} reaches 29.0 pass@8, compared with 20.3 for UI-TARS-1.5 and 21.7 for noisy ARPO from the base model.
    \item \textbf{Data quality:} A 30\%-data \textsc{ENVS} variant reaches 27.0 pass@8, matching or exceeding ARPO from the base model.
    \item \textbf{Robustness trade-off:} Clean collection performs best on clean \textsc{OSWorld}, while noisy collection slightly improves \textsc{OSWorld-Noisy} and better preserves auxiliary capabilities, e.g., \textsc{OSWorld-G Refusal} (16.7 vs. 1.9).
\end{itemize}

\begin{figure}[H]
\centering
\vspace{-1em}
\includegraphics[
  width=\linewidth,
  clip
]{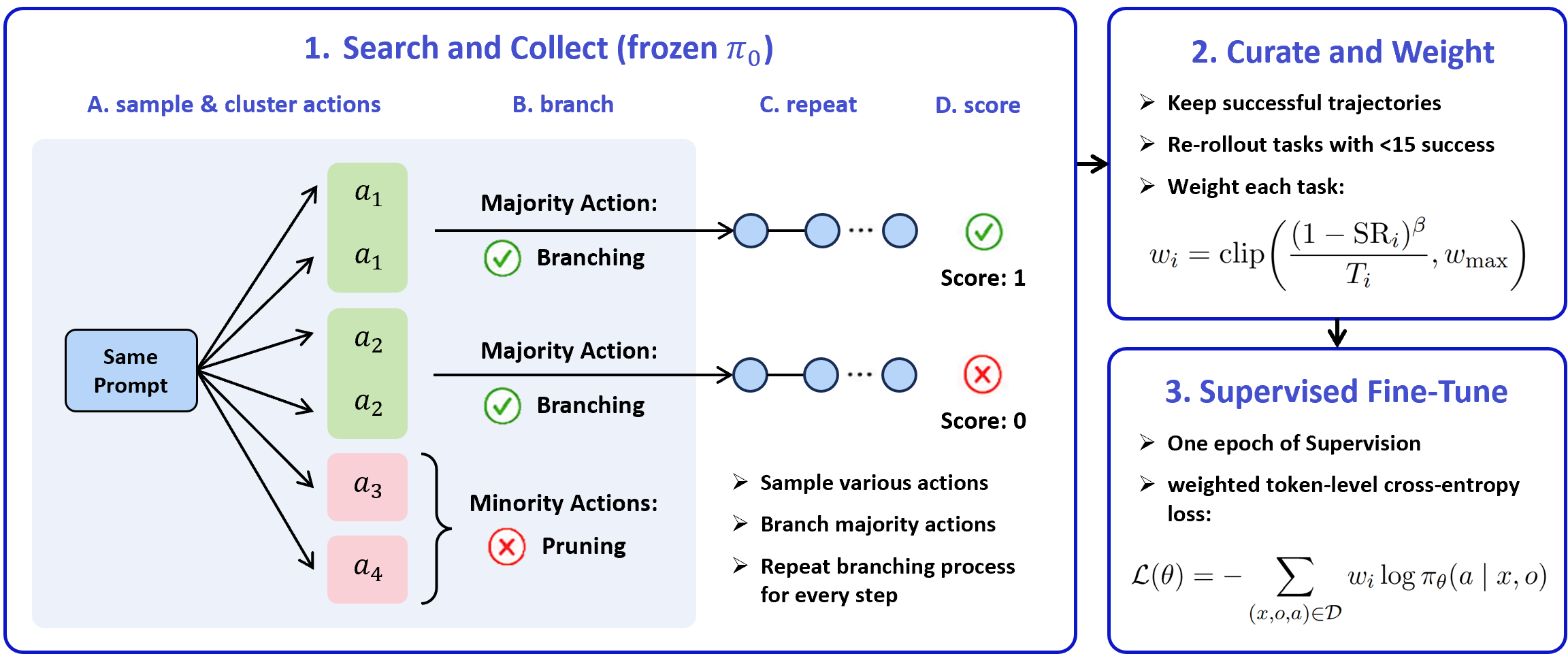}
\caption{\textbf{\textsc{ENVS} pipeline overview}.
\textsc{ENVS} uses environment-native tree search to collect verified successful trajectories from \textsc{OSWorld}, curates them through filtering, weighting, and deduplication, and trains the agent with one-epoch SFT before evaluation on clean and noisy benchmarks.}
\label{fig:pipeline}
\end{figure}

\section{Related Work}
\paragraph{GUI agents and executable environments.}
Recent GUI-agent work studies how vision-language models map screenshots and instructions to executable actions across desktop, web, and mobile interfaces. \textsc{OSWorld} provides real-desktop execution-based evaluation~\citep{xie2024osworld}, while WebArena, VisualWebArena, AndroidWorld, and Mind2Web study related web and mobile settings~\citep{zhou2023webarena,koh2024visualwebarena,rawles2024androidworld,deng2024mind2web}. Model-side systems such as SeeClick, CogAgent, ScreenAgent, UI-TARS, OS-Atlas, and Aguvis improve GUI grounding and action prediction~\citep{cheng2024seeclick,hong2023cogagent,zhang2024screenagent,qin2025uitars,wu2024osatlas,zhang2025aguvis}. Our work is complementary: we use executable GUI environments to construct verified training trajectories rather than only to evaluate or collect online rollouts.

\paragraph{RL, search, and verified supervision for agents.}
Reinforcement learning with verifiable rewards is widely used for language-model training~\citep{schulman2017ppo,shao2024deepseekmath,deepseek2025r1,yu2025dapo}. In GUI and agent settings, ARPO adapts GRPO-style optimization with replay~\citep{lu2025arpo_gui}, Agentic ARPO studies adaptive rollouts for tool-using agents~\citep{dong2025agentic_arpo}, ToolRL and RAGEN train agents with environment or tool feedback~\citep{qian2025toolrl,wang2025ragen}, and SEEA-R1 combines MCTS-style exploration with online policy optimization~\citep{tian2025seea}. Tree search has also been used to turn sparse feedback into stronger policies, from MCTS/UCT and expert iteration to AlphaGo, AlphaZero, and MuZero~\citep{coulom2006mcts,kocsis2006uct,anthony2017expert,silver2016alphago,silver2017alphazero,schrittwieser2020muzero}, and recent language-agent methods explore multiple reasoning or action paths before committing~\citep{yao2023treeofthoughts,koh2024tree_search_lm_agents,putta2024agentq}. \textsc{ENVS} differs by using search only for training-time discovery of environment-verified demonstrations, followed by globally balanced supervised learning rather than online RL updates or deployment-time planning.

\paragraph{Data selection, imitation, verification, and robustness.}
\textsc{ENVS} is related to curriculum and data-selection methods such as CLPO, which emphasize informative examples during policy optimization~\citep{zhang2025clpo}, as well as imitation learning~\citep{pomerleau1989alvinn,ross2011dagger} and verifier/process-supervision methods~\citep{cobbe2021verifiers,lightman2023process,gulcehre2023rest}. Its supervision, however, is generated by live GUI search and filtered by terminal execution success. \textsc{OSWorld-Noisy} connects to work on curriculum learning, domain randomization, procedural generation, prioritized level replay, unsupervised environment design, and robust RL~\citep{bengio2009curriculum,tobin2017domain_randomization,cobbe2020procgen,jiang2021plr,dennis2020paired,parkerholder2022accel,pinto2017robust_adversary,zhang2020robust_rl,slaoui2020robust_domain_randomization}, but instantiates these ideas as recoverable desktop interruptions such as popups, overlays, focus shifts, and dialogs in executable GUI environments.

\section{ENVS: Environment-Native Verified Search}

ENVS is an SFT-based, training-time pipeline for \textsc{OSWorld} GUI agents. It searches in live desktop VMs, verifies successful leaves, and converts them into balanced action-level supervision. This separation lets ENVS inspect the full discovered dataset before training, cap overrepresented tasks, normalize trajectory length, and control gradient allocation across tasks.

A \emph{step} is one action index in a trajectory. A \emph{node} is an environment branch: an executed action prefix and the desktop state reached by that prefix. An \emph{alive node} is a branch still being expanded. When ENVS branches, the parent follows one action bucket, while child VMs replay the parent prefix and execute alternative behaviorally distinct actions.

\subsection{Behavioral-disagreement-gated search}

At each live node, ENVS samples $K$ candidate next actions from a frozen policy and reduces each to a coarse behavior fingerprint: action type plus a discretization of its argument (quantized coordinates for clicks and drags, normalized text for typing, direction for scrolls, action type alone for zero-argument actions). Fingerprints are a cheap proxy, not a semantic equivalence --- nearby clicks may hit different widgets, and distant clicks may hit the same large button. Candidates are bucketed by fingerprint, and the buckets are ranked by agreement: the number of sampled actions that fall in each bucket. The $k$ highest-agreement buckets, where $k$ is the per-step branch budget, are the \emph{majority} actions and are explored; all lower-ranked buckets are \emph{minority} actions and are pruned. If only one bucket is populated, the node continues along it; otherwise the single highest-agreement action is carried forward on the parent VM (so no VM is spent re-reaching the current state), and each remaining majority action spawns a child VM that replays the prefix and executes that divergent behavior. VM budget thus tracks behavioral divergence, not token noise.

\subsection{VM-bounded search in live desktop environments}
ENVS borrows the branching structure of tree search but not its standard machinery. Classical MCTS variants assume cheap, repeated simulations from abstract states and rely on UCB/PUCT selection with learned values~\citep{coulom2006mcts,kocsis2006uct,silver2016alphago,silver2017alphazero,schrittwieser2020muzero}. In OSWorld, each branch is a live VM trajectory (real clicks, typing, application latency, focus state, brittle desktop transitions) and revisiting an abstract state is not free.

We therefore replace adaptive selection with explicit cost gates: a per-task branch budget, a per-step branch cap, late-step deferral, and a hard step cutoff. ENVS uses no learned value function, no UCB/PUCT, and no edge visit statistics. Its job is training-time trajectory discovery: finding executable paths
that succeed under the \textsc{OSWorld} evaluator and converting them to supervised data.

\subsection{Environment verification and successful-leaf filtering}
After search, the OSWorld task oracle assigns each leaf trajectory a binary reward $R(\tau) \in \{0,1\}$. Failed trajectories are discarded; successful trajectories are decomposed into per-step supervised examples,

\[\mathcal{D} = \{(x, o_{\le t}, a_t) : R(\tau)=1,\; (o_{\le t}, a_t) \in \tau\},\]

where $x$ is the task instruction, $o_{\le t}$ the observation history, and $a_t$ the executed action. A single successful long-horizon trajectory thus yields many supervised targets, converting sparse terminal success into dense action-level supervision.

\subsection{Global balancing of action-level supervision}
Search produces an imbalanced dataset: easy tasks yield many successful leaves, long trajectories contribute many step examples, while hard tasks may yield few or none. Trained uniformly, the gradient is dominated by easy tasks and long trajectories.

ENVS rebalances after collection. Let $\mathcal{D}_i$ be the successful step examples for task $i$, $T_i = |\mathcal{D}_i|$, and $\mathrm{SR}_i$ its search success rate. The training objective is

\[
\begin{aligned}
\mathcal{L}(\theta)
&=
\sum_i
\sum_{(x,o_{\le t},a_t)\in \mathcal{D}_i}
w_i
\left[-\log \pi_\theta(a_t \mid x,o_{\le t},a_{<t})\right], \\
w_i
&=
\mathrm{clip}\!\left(
\frac{(1-\mathrm{SR}_i)^\beta}{T_i},
w_{\max}
\right).
\end{aligned}
\]
$(1-\mathrm{SR}_i)^\beta$ shifts gradient mass to hard tasks; $1/T_i$ normalizes trajectory; $w_{\max}$ caps rare-success outliers. Smaller $\beta$ ensures uniform weighting; larger $\beta$ emphasizes hard tasks more aggressively.

This dataset-level rebalancing is the structural payoff of decoupling search from optimization: with the full verified dataset in hand before training, ENVS can flatten gradient allocation globally. Online RL gradients are determined by whatever rollouts the current policy produces, with no comparable handle on per-task gradient allocation.

\section{\textsc{OSWorld-Noisy}: Held-out Desktop Perturbations}

Standard \textsc{OSWorld} evaluation assumes a quiescent interaction
loop in which the agent is the only active process on the desktop. In
deployment this assumption rarely holds: focus changes, notifications,
modal dialogs, and concurrent background activity routinely alter the
visible state during task execution. \textsc{OSWorld-Noisy} measures
whether GUI agents can complete the original \textsc{OSWorld} tasks
under such interruptions. We model the perturbation source as a
concurrent benign user who performs short, observable actions in
non-target applications while the agent works on its task; the
resulting friction is recoverable and is layered over the unmodified
task and evaluator.

\subsection{Perturbation templates}
\label{sec:noise:library}

\textsc{OSWorld-Noisy} draws from a library of $153$ runtime noise
generators spanning three categories: concurrent human-task sessions
in non-target applications (e.g., note-taking in \texttt{gedit},
folder navigation in \texttt{nautilus}, or terminal activity); ambient
interruptions (notifications, dialogs, browser prompts, toast
overlays); and rare accidental target interference (partial occlusion,
window shove, resize). Each generator returns a bash command that
executes a short, observable interaction inside the \textsc{OSWorld}
container. A held-out subset of $24$ generators is reserved
exclusively for evaluation; the remaining $129$ are available for
noisy training and collection. The full catalog and held-out split
are provided in Appendix~\ref{app:noise}.

All perturbations satisfy three non-sabotage constraints: (i) noise
does not deliver keystrokes or clicks to windows it has not itself
opened and focused; (ii) noise does not touch files or paths
referenced by the task evaluator; (iii) every perturbation is
recoverable within a small number of standard agent actions. Under
these constraints, \textsc{OSWorld-Noisy} is a robustness setting
rather than an impossibility benchmark: interruptions may distract or
delay the agent but cannot render any task unsolvable.

\subsection{Schedule sampler}

A schedule specifies, for each rollout, the firing time of each
perturbation, the bash command to execute, and metadata consumed by
the curriculum. During training, perturbation count and difficulty
scale with the task's recent success rate: hard tasks receive little
or no noise, while higher-success tasks receive more frequent or
higher-cost events. This protects tasks the policy cannot yet solve
and introduces interruption gradually as performance improves.

Evaluation uses a fixed protocol. Each task receives exactly one
held-out perturbation with deterministic timing, and the schedule is
shared across models via common random numbers. Clean and noisy
evaluations are therefore directly comparable, and cross-model
differences cannot be attributed to random variation in noise
difficulty.

\subsection{Train/evaluation split}

The $24$ held-out generators are reserved for evaluation; the
underlying \textsc{OSWorld} tasks are unchanged. \textsc{OSWorld-Noisy}
therefore measures recovery under unseen interruptions on the same
task set, not performance on a different task set. This split follows
the standard environment-generalization principle that robustness be
evaluated on variations related to, but distinct from, those seen
during training~\citep{cobbe2020procgen,jiang2021plr,dennis2020paired,parkerholder2022accel}.
In our setting, the variation is a recoverable GUI interruption
rather than a procedurally generated level or simulated physics
parameter.

\subsection{Injection sites}

The same library supports three uses. During ENVS collection, a
fraction of search branches receive noise while the remainder serve
as clean controls, producing a mix of task-solving and recovery
trajectories. During on-policy RL rollout collection (e.g., ARPO), a
single schedule is shared across the rollout group for a task,
reducing environment-driven variance within the group. At evaluation,
schedules are precomputed from the held-out templates and executed
verbatim, ensuring that \textsc{OSWorld-Noisy} is reproducible across
models and runs.

\section{Experiments and Results}
\label{sec:experiments}

\subsection{Experimental protocol}
\label{sec:exp_protocol}

We evaluate on a fixed 300-task \textsc{OSWorld} pool consisting of
86 trainable tasks and 214 held-out tasks. The trainable subset follows
the ARPO protocol: it contains tasks on which the base UI-TARS-1.5-7B
policy occasionally succeeds, ensuring non-zero reward variation for
group-relative online RL. The held-out tasks are never used for ENVS
collection or ARPO training. We evaluate each method on clean
\textsc{OSWorld} and on \textsc{OSWorld-Noisy}, which uses the same
300 tasks with one held-out recoverable perturbation per task.

All policies are initialized from UI-TARS-1.5-7B. We compare ENVS with
the base policy and ARPO, a GRPO-based online-RL baseline with replay
over successful trajectories. The primary metric is pass@8: a task is
solved if at least one of eight rollouts sampled at temperature
$T=1.0$ succeeds within the 15-step horizon. All methods share the same
evaluator, horizon, temperature, and sampling protocol. We report GPU
hours because live VM execution dominates training cost; ARPO collects
fresh on-policy rollouts during training, while ENVS collects verified
trajectories once and reuses them for SFT variants.

\subsection{Main results on \textsc{OSWorld} and \textsc{OSWorld-Noisy}}

All methods start from UI-TARS-1.5-7B and are evaluated with the same
15-step pass@8 protocol on the 300-task pool. ENVS is the strongest
method in both matched conditions (Table~\ref{tab:main-results}). On
clean \textsc{OSWorld}, ENVS reaches 30.3\% overall pass@8, improving
over the base model by 7.6 pp and ARPO-clean by 3.6 pp. The gain appears
on both trained tasks (73.3\% to 87.2\%) and held-out tasks (2.3\% to
7.5\%).

On \textsc{OSWorld-Noisy}, ENVS reaches 29.0\%, compared with 20.3\%
for the base model and 21.7\% for ARPO-noisy. The improvement is mainly
concentrated on the trained subset: ENVS-noisy reaches 88.4\%, while
ARPO-noisy reaches 62.8\% and falls below the noisy base model. On
held-out tasks, ENVS-noisy and ARPO-noisy both reach 5.1\%, so the noisy
gain reflects stronger use of verified trainable-task supervision rather
than broad held-out transfer.

\begin{table}[H]
  \caption{Main results on \textsc{OSWorld} and \textsc{OSWorld-Noisy}. We
  report pass@8 success rate (\%) on the trained 86-task subset, the
  held-out subset, and overall, with absolute pp gain over the base
  policy in parentheses. The trained subset is selected following the
  ARPO protocol as the tasks on which the base policy occasionally
  succeeds. Each training method is evaluated only on its matched eval
  condition. Bold marks the column maximum within each block.}
  \label{tab:main-results}
  \centering
  \small
  \begin{tabular}{llccc}
    \toprule
    Model           & Train data         & Trained ($86$)            & Held-out                 & Overall                    \\
    \midrule
    \multicolumn{5}{l}{\textit{\textsc{OSWorld} (clean)}} \\
    UI-TARS-1.5-7B  & ---                & $73.3$                    & $2.3$                    & $22.7$                     \\
    + ARPO          & clean rollouts     & $81.4\,(+8.1)$            & $4.7\,(+2.4)$            & $26.7\,(+4.0)$             \\
    + ENVS          & clean trajectories & $\mathbf{87.2\,(+13.9)}$  & $\mathbf{7.5\,(+5.2)}$   & $\mathbf{30.3\,(+7.6)}$    \\
    \midrule
    \multicolumn{5}{l}{\textit{\textsc{OSWorld-Noisy}}} \\
    UI-TARS-1.5-7B  & ---                & $66.3$                    & $1.9$                    & $20.3$                     \\
    + ARPO          & noisy rollouts     & $62.8\,(-3.5)$            & $5.1\,(+3.2)$            & $21.7\,(+1.4)$             \\
    + ENVS          & noisy trajectories & $\mathbf{88.4\,(+22.1)}$  & $\mathbf{5.1\,(+3.2)}$   & $\mathbf{29.0\,(+8.7)}$    \\
    \bottomrule
  \end{tabular}
\end{table}

\subsection{Compute efficiency}
\label{sec:compute}

ENVS uses 138--153 total GPU-hours, compared with 184--192 for matched
ARPO baselines (Table~\ref{tab:compute}).
The saving comes from decoupling collection from optimization: ENVS
collects verified trajectories once and trains only successful trajectories with one-epoch SFT,
whereas ARPO collects fresh on-policy rollouts inside each training run, updating policy with all trajectories.

\begin{table}[H]
  \caption{Per-method compute decomposition. ENVS rows split SFT
  GPU-hours from offline trajectory-collection GPU-hours; ARPO fuses
  rollout collection and policy updates on the training cluster (Collect
  column dashed). Bold marks the column minimum within each block.}
  \label{tab:compute}
  \centering
  \small
  \begin{tabular}{llccc}
    \toprule
    Model           & Method           & Train GPU-h & Collect GPU-h & Total GPU-h                 \\
    \midrule
    \multicolumn{5}{l}{\textit{Clean-trained}} \\
    UI-TARS-1.5-7B  & + ARPO           & $184$       & ---           & $184$                       \\
    UI-TARS-1.5-7B  & + ENVS           & $31$        & $107$         & $\mathbf{138}$              \\
    \midrule
    \multicolumn{5}{l}{\textit{Noisy-trained}} \\
    UI-TARS-1.5-7B  & + ARPO           & $192$       & ---           & $192$                       \\
    UI-TARS-1.5-7B  & + ENVS           & $32$        & $121$         & $\mathbf{153}$              \\
    \bottomrule
  \end{tabular}

  \vspace{0.9em}
  \begin{minipage}{0.92\linewidth}
  \footnotesize
  \emph{Note.} ARPO is an online policy-optimization method: rollout collection occurs inside the training loop on the same training GPUs, so collection cost is included in Train GPU-h and is not reported as a separate offline phase. ENVS decouples the phases, so trajectory collection and SFT training are reported separately.
  \end{minipage}
\end{table}

\subsection{Data efficiency}
\begin{figure}[H]
\centering
\includegraphics[width=0.45\linewidth]{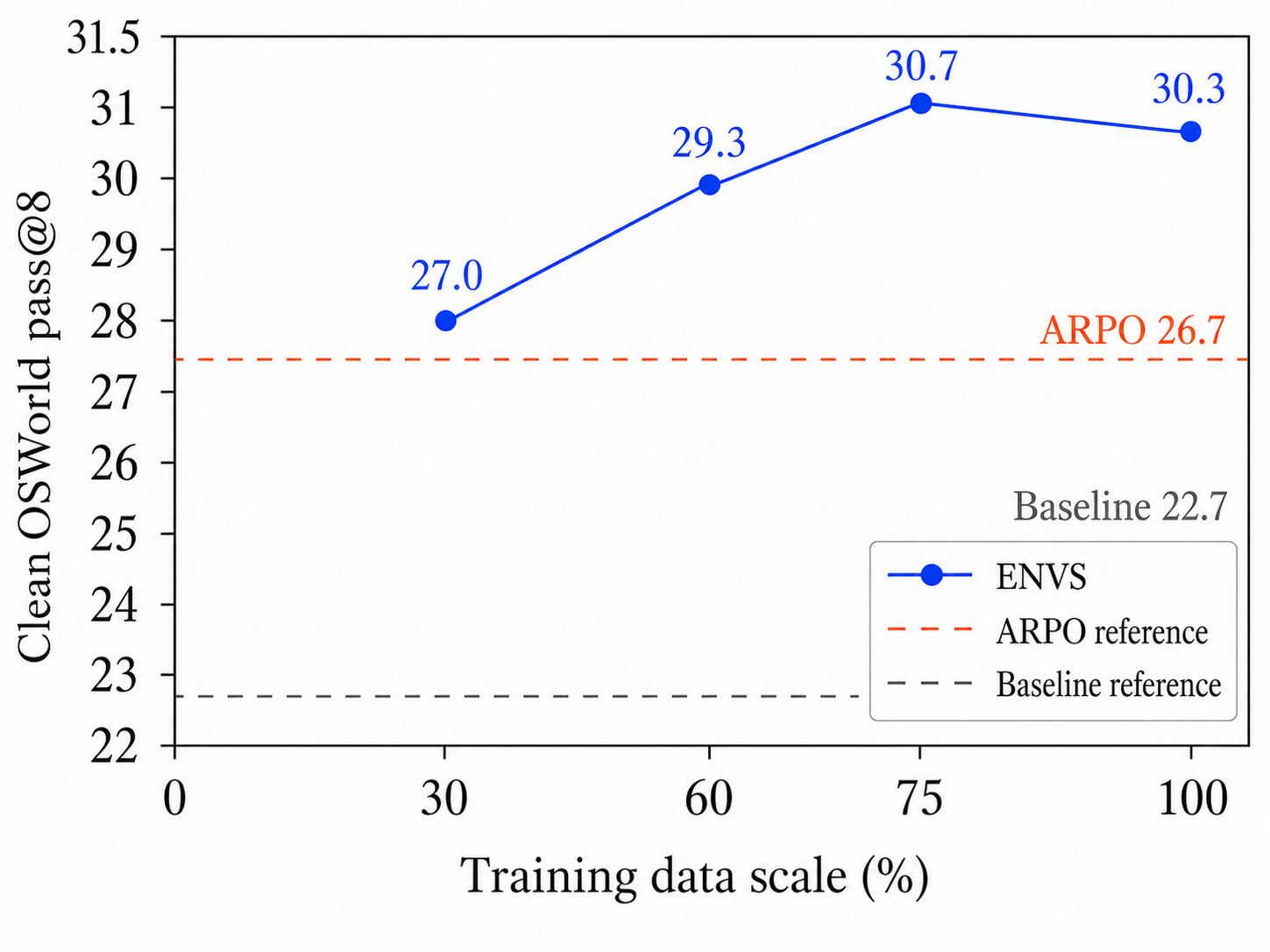}
\caption{Clean \textsc{OSWorld} pass@8 as a function of ENVS training
data volume. The 30\% subset already matches ARPO-clean, while gains
saturate near the full dataset.}
\label{fig:data-efficiency}
\end{figure}

ENVS remains competitive with substantially less data. Using only 30\%
of collected trajectories, it reaches 27.0\% pass@8, matching ARPO-clean
at 26.7\% (Figure~\ref{fig:data-efficiency}). Performance increases to
30.7\% at 75\% data and remains essentially saturated at 100\%
(30.3\%). This suggests that verified search improves supervision
quality, not only supervision volume.

\subsection{Clean versus noisy trajectory collection}
\label{sec:clean-vs-noisy}

Table~\ref{tab:clean-vs-noisy} evaluates each trained policy on both
clean and noisy test conditions. Under the matched ARPO configuration,
noisy online rollouts degrade performance: ARPO-noisy reaches only
22.3\% on clean \textsc{OSWorld} and 21.7\% on
\textsc{OSWorld-Noisy}, close to the base model and below ARPO-clean's
noisy transfer score of 23.7\%. Its trained-subset performance also
drops below the noisy base model (66.3\% to 62.8\%;
Table~\ref{tab:main-results}).

ENVS is more stable across conditions. ENVS-clean scores 30.3\% on
clean evaluation and 28.3\% under noise, while ENVS-noisy scores 29.0\%
on both. Noisy collection therefore does not improve task completion
over clean collection, but ENVS avoids the degradation observed in noisy
online RL.

\begin{table}[H]
  \caption{Cross-evaluation of every trained policy on both the clean
  and noisy $300$-task pools. Values are pass@8 in percent. Bold marks
  the column maximum. ARPO-noisy is the weakest learned policy and is
  matched by ARPO-clean's cross-condition transfer; ENVS-noisy reaches
  parity with ENVS-clean on both conditions.}
  \label{tab:clean-vs-noisy}
  \centering
  \small
  \begin{tabular}{lcc}
    \toprule
    Variant            & Clean OSWorld   & \textsc{OSWorld-Noisy}    \\
    \midrule
    UI-TARS-1.5-7B     & $22.7$          & $20.3$                    \\
    + ENVS, clean      & $\mathbf{30.3}$ & $28.3$                    \\
    + ENVS, noisy      & $29.0$          & $\mathbf{29.0}$           \\
    + ARPO, clean      & $26.7$          & $23.7$                    \\
    + ARPO, noisy      & $22.3$          & $21.7$           \\
    \bottomrule
  \end{tabular}
\end{table}

\subsection{Auxiliary visual-reasoning capability retention}
\label{sec:diversity}

Table~\ref{tab:aux} evaluates whether GUI training preserves broader
visual-reasoning abilities. Noisy ENVS matches or exceeds clean ENVS on
all twelve benchmarks. The largest difference is OSWorld-G refusal:
clean ENVS drops from 14.8\% to 1.9\%, while noisy ENVS reaches 16.7\%.
Noisy ENVS also improves over clean ENVS on BLINK Functional
Correspondence (23.1\% to 26.2\%) and BLINK IQ Test (27.3\% to 30.0\%).

These results indicate a trade-off. Clean ENVS gives the best clean
task-completion score, while noisy ENVS better preserves auxiliary
visual-reasoning behavior. This suggests that recoverable interruptions
provide useful supervision for perception, waiting, and recovery, even
when they do not improve clean task completion.

\begin{table}[H]
  \caption{Auxiliary visual-reasoning performance, grouped by capability
  channel. All scores are canonical benchmark accuracy (\%) under
  VLMEvalKit deterministic decoding; no runtime perturbation is applied
  at evaluation. Bold marks the best of the three per row. Noisy ENVS
  matches or exceeds clean ENVS on every benchmark.}
  \label{tab:aux}
  \centering
  \small
  \begin{tabular}{lccc}
    \toprule
    Benchmark                                 & Base            & Clean ENVS & Noisy ENVS      \\
    \midrule
    \multicolumn{4}{l}{\textit{Refusal calibration}} \\
    \quad OSWorld-G refusal                   & $14.8$          & $1.9$      & $\mathbf{16.7}$ \\
    \midrule
    \multicolumn{4}{l}{\textit{Vision-indispensable reasoning}} \\
    \quad MathVerse-VO multi-choice           & $\mathbf{78.0}$ & $66.7$     & $71.1$          \\
    \quad MMStar math                         & $\mathbf{52.8}$ & $47.6$     & $50.8$          \\
    \quad MMStar logical reasoning            & $\mathbf{60.0}$ & $56.0$     & $\mathbf{60.0}$ \\
    \quad MMStar science \& technology        & $\mathbf{38.4}$ & $36.4$     & $37.6$          \\
    \midrule
    \multicolumn{4}{l}{\textit{Visual-prior override}} \\
    \quad HallusionBench VD-figure            & $68.8$          & $67.5$     & $\mathbf{70.0}$ \\
    \quad HallusionBench VD-illusion          & $\mathbf{63.2}$ & $54.2$     & $56.3$          \\
    \midrule
    \multicolumn{4}{l}{\textit{Cross-image differentiation}} \\
    \quad MMVP per-pair                       & $44.7$          & $44.7$     & $\mathbf{46.7}$ \\
    \quad BLINK Semantic Correspondence       & $33.1$          & $32.4$     & $\mathbf{33.8}$ \\
    \midrule
    \multicolumn{4}{l}{\textit{Abstract reasoning}} \\
    \quad BLINK Functional Correspondence     & $23.1$          & $23.1$     & $\mathbf{26.2}$ \\
    \quad BLINK IQ Test                       & $27.3$          & $27.3$     & $\mathbf{30.0}$ \\
    \midrule
    \multicolumn{4}{l}{\textit{Spatial reasoning}} \\
    \quad BLINK Spatial Relation              & $87.4$          & $87.4$     & $\mathbf{88.1}$ \\
    \bottomrule
  \end{tabular}
\end{table}

\section{Conclusion}

Realistic GUI control brings embodied-AI challenges into software
environments: agents must act through perception and action in stateful
desktops where observations change, actions have delayed effects, and
interruptions can occur during task execution. \textsc{ENVS} addresses
this setting by turning sparse terminal feedback from live desktop
environments into verified, balanced action-level supervision. It
searches in \textsc{OSWorld} VMs, branches over behaviorally distinct
GUI actions, filters successful leaves with the environment oracle, and
trains from globally balanced trajectories.

Across completed \textsc{OSWorld} experiments, \textsc{ENVS}
outperforms ARPO-style online RL on clean and noisy evaluation while
using less compute. The reduced-data result shows that a
small set of rich, diverse, environment-verified trajectories can reach
online-RL-level performance, suggesting that carefully constructed SFT
data can recover much of the benefit of RL without repeated on-policy
VM rollouts. \textsc{OSWorld-Noisy} further shows that dynamic
perturbations are not only a robustness test, but also a source of
supervision for refocusing, waiting, recovery, and careful perception.

More broadly, environment-native training can make agent
training cheaper and more reusable by amortizing expensive environment
interaction into verified trajectory datasets. At the same time,
stronger GUI-control agents raise risks from unintended actions,
unauthorized software control, and harmful automation. Our experiments
are restricted to benchmark VMs with recoverable perturbations and
task-specific evaluators; future work should pair environment-native
training with permission boundaries, auditing, human oversight, and
deployment-time safety constraints.

\clearpage

{\small
\bibliographystyle{plainnat}
\bibliography{references}
}

\clearpage
\appendix

\section{Preliminaries}

\subsection{GUI agents as partially observed control}

An \textsc{OSWorld} task specifies a natural-language instruction \(x\) and an initial desktop state. At each step \(t\), the agent observes a context \(o_t\) consisting of the current screenshot and interaction history, then emits an executable GUI action \(a_t\), such as clicking, typing, scrolling, pressing a hotkey, waiting, or terminating. Since the full desktop state is only partially observed through screenshots and actions can have delayed effects, we write the policy as history-conditioned:
\[
    \pi_\theta(a_t \mid x, o_{\leq t}, a_{<t}).
\]
A trajectory is a sequence
\[
    \tau = (o_0, a_0, o_1, a_1, \ldots, o_H, a_H)
\]
executed in a live virtual machine. Because actions modify desktop state, token-distinct actions can induce the same GUI behavior, while superficially similar actions can lead to different future states.

\subsection{Sparse terminal verification in GUI environments}

\textsc{OSWorld} provides task-specific evaluators that judge whether a completed trajectory satisfies the instruction. For ENVS, we use this evaluator as a binary terminal verifier,
\[
    R_{\mathrm{env}}(\tau) \in \{0,1\}.
\]
This signal is sparse: most intermediate states do not carry direct supervision, and a terminal failure does not identify which earlier action caused the failure. Online RL methods use related terminal reward signals to update the current policy from its own rollouts. Depending on the method, these rewards may include shaping terms or penalties for invalid outputs rather than being strictly binary, but they remain sparse because they are assigned primarily at the trajectory or final-output level \citep{dong2025agentic_arpo, lu2025arpo_gui}. ENVS uses the environment signal differently: it treats \(R_{\mathrm{env}}(\tau)\) as a verifier for search-generated trajectories, filters successful leaves, and then trains from the resulting verified supervision.

\section{Additional implementation details}

\subsection{Noise template catalog}
\label{app:noise}

\textsc{OSWorld-Noisy} contains 153 runtime noise generators. A held-out subset of 24 generators is used only for evaluation. Table~\ref{tab:noise-taxonomy-full} gives the expanded taxonomy used by the implementation. Counts refer to generator entries; a single generator can produce many concrete events through internal randomization.

\begin{table}[h]
\centering
\caption{Expanded taxonomy of \textsc{OSWorld-Noisy} perturbations. Counts are generator entries; one generator can produce many concrete events through internal randomization.}
\label{tab:noise-taxonomy-full}
\scriptsize
\begin{tabular}{p{0.18\linewidth} p{0.07\linewidth} p{0.43\linewidth} p{0.24\linewidth}}
\toprule
Tier (count) & Cost & Representative perturbations & Recovery behavior \\
\midrule
T1: concurrent human-task session (45)
& 0--1
& \texttt{gedit} notes and drafts; \texttt{gnome-terminal} commands; \texttt{nautilus} folder browsing; settings panels; window switching and dragging.
& Refocus the target application. Noise does not deliver input to the target window. \\
\midrule
T2: ambient background and interruption (99)
& 0--3
& Notifications, update/backup dialogs, fake downloads, browser prompts, permission dialogs, OS toasts, banners, modal overlays, compositional notification/modal events.
& Dismiss, press \texttt{Esc}, ignore, wait, or dismiss components in sequence. \\
\midrule
T3: accidental target interference (9)
& 1--2
& Target-window shove, shrink, partial overlap, cover-from-above, transient network/proxy prompts.
& Refocus, move, resize, or wait for auto-recovery. \\
\bottomrule
\end{tabular}
\end{table}

\subsection{Noise scheduling}
\label{app:noise-scheduling}

During training, the number and difficulty of perturbations depend on task success. Hard tasks receive little or no noise; easier tasks receive more frequent and higher-cost events. At evaluation, each task receives exactly one held-out perturbation event. The evaluation schedule is deterministic, so every model sees the same interruption for the same task.

\paragraph{Fire count.}
\[
N(\hat{s}) =
\begin{cases}
0, & \hat{s}<0.10, \\
1, & 0.10 \le \hat{s} < 0.25, \\
\mathrm{Unif}\{1,2,3\}, & 0.25 \le \hat{s} < 0.50, \\
\mathrm{Unif}\{3,4,5\}, & \hat{s} \ge 0.50.
\end{cases}
\]
At evaluation, \(N\equiv 1\).

\paragraph{Step placement.}
Fire steps are bucket-spaced across the rollout horizon. Setup steps are protected, and each scheduled event leaves enough remaining steps for recovery and task completion.

\paragraph{Determinism.}
Training and collection schedules are seeded by task and rollout identifiers. Evaluation schedules are precomputed from the held-out generator set and executed verbatim across all models.

\section{Evaluation pool construction}
\label{app:eval-pool}

We build the 300-task evaluation pool by applying two filters to the
public \textsc{OSWorld} test suite of $368$ tasks across $10$ application categories.

\paragraph{Filter~1: soft-reset stability ($-13$).}
A task is included only if its evaluator returns a deterministic outcome
when the desktop is reset via OSWorld's soft-reset path (revert VM to a
named snapshot rather than a full re-provision). Tasks whose evaluators
fail intermittently under soft reset --- typically because they depend on
filesystem state that the snapshot does not restore --- are removed. This
filter excludes $13$ tasks; the surviving set is the
\texttt{test\_all\_softreset.json} list of $355$ tasks.

\paragraph{Filter~2: proxy-free reachability ($-55$).}
A task is included only if every URL its trajectory reaches can be served
without an authenticated outbound HTTP proxy. Browser tasks that depend on
gated services (logged-in social sites, region-locked endpoints, accounts
we cannot anonymously instantiate) are removed. This filter excludes a
further $55$ tasks, leaving the $300$-task evaluation pool used in this
paper.

The two filters together exclude $68$ of $368$ tasks. The exclusions are
concentrated in two app categories whose tasks are predominantly browser-
or web-mediated (Table~\ref{tab:eval-pool-breakdown}); GIMP, LibreOffice
Calc, LibreOffice Writer, and VLC are unaffected.

\begin{table}[h]
\centering
\caption{Per-app composition of the evaluation pool. The 300-task pool
preserves all $10$ \textsc{OSWorld} application categories. Exclusions are
concentrated in \texttt{chrome} and \texttt{multi\_apps}; both are
dominated by browser- and network-mediated tasks affected by Filter~2.}
\label{tab:eval-pool-breakdown}
\small
\begin{tabular}{lrrr}
\toprule
Application & Full (368) & Eval pool (300) & Excluded \\
\midrule
\texttt{chrome}              & 46  & 16  & 30 \\
\texttt{gimp}                & 26  & 26  & 0  \\
\texttt{libreoffice\_calc}    & 47  & 47  & 0  \\
\texttt{libreoffice\_impress} & 47  & 46  & 1  \\
\texttt{libreoffice\_writer}  & 23  & 23  & 0  \\
\texttt{multi\_apps}          & 101 & 72  & 29 \\
\texttt{os}                  & 24  & 19  & 5  \\
\texttt{thunderbird}         & 15  & 14  & 1  \\
\texttt{vlc}                 & 17  & 17  & 0  \\
\texttt{vs\_code}             & 22  & 20  & 2  \\
\midrule
Total                        & 368 & 300 & 68 \\
\bottomrule
\end{tabular}
\end{table}

\paragraph{Train / eval relationship.}
The 86-task trainable subset used for ENVS collection and ARPO training
is a strict subset of the 300-task evaluation pool ($86 \subset 300$).
Pass@8 numbers reported in Section~\ref{sec:experiments} therefore include both trained-on
tasks and 214 held-out tasks; the train / held-out split is broken out
in Table~\ref{tab:main-results}.

\paragraph{Exact list.}
The $300$ retained task IDs and their per-app grouping are shipped
verbatim in the supplementary archive as
\path{evaluation_examples/test_all_300tasks_noproxy_softreset_clean.json}.
The $355$-task soft-reset-filtered intermediate set is provided as
\path{evaluation_examples/test_all_softreset.json} for readers who wish to
reproduce the two filter stages independently.

\section{Structural Advantages of \textsc{ENVS}}
\label{app:envs-vs-rl}

The empirical results in Tables~\ref{tab:main-results}--\ref{tab:clean-vs-noisy}
show that \textsc{ENVS} outperforms ARPO on long-horizon, sparse-reward
GUI control under both clean and noisy evaluation. This appendix
identifies four structural properties of online policy gradient that
account for the gap and shows that each is removed by construction in
\textsc{ENVS}: a binomial degeneracy in per-task gradient allocation,
monotone contraction of action entropy under positive-covariance updates,
non-convexity of the return objective combined with unbounded equilibrium
drift, and the inability to reuse rollouts across training variants.
Notation follows the main body throughout: $\pi_\theta$ for the trained
policy, $\pi_0$ for UI-TARS-1.5-7B, $\mathcal{D}_i$ for the supervised
step set on task $i$ with $T_i = |\mathcal{D}_i|$, and $\mathrm{SR}_i$
for the empirical search success rate on task $i$. We additionally write
$q(\cdot \mid s)$ for the empirical action distribution at state $s$
implied by $\bigcup_i \mathcal{D}_i$, and $H[\pi(\cdot \mid s)]$ for the
Shannon entropy of $\pi$ at state $s$.

\subsection{Gradient Degeneracy}
\label{app:rl:gradient}

Under independent Bernoulli rollouts with per-trajectory success
probability $p_i$ on task $i$, the number of successes $k$ in a group of
size $G$ follows $\mathrm{Bin}(G, p_i)$. The probability that the group
contains both at least one success and at least one failure is therefore
\[
    \mathrm{Pr}[\text{non-degenerate}]
    \;=\; 1 - \mathrm{Pr}[k = 0] - \mathrm{Pr}[k = G]
    \;=\; 1 - (1-p_i)^G - p_i^G.
\]
The (uncorrected) sample variance of $G$ Bernoulli outcomes with $k$
successes is
\[
    \hat\sigma_r^2 \;=\; \frac{1}{G}\sum_{j=1}^{G} (r_j - \bar r)^2
    \;=\; \bar r\,(1 - \bar r)
    \;=\; \frac{k}{G}\!\left(1 - \frac{k}{G}\right),
\]
which equals zero when $k \in \{0, G\}$. The implementation guards the
resulting division in the standardised advantage by an $\varepsilon$,
sending all advantages on a degenerate group to approximately zero;
direct calculation on a non-degenerate group gives $\sum_{j=1}^G \hat
A_j^2 = G$. Per-task expected gradient mass therefore factorises as the
non-degenerate probability above multiplied by the conditional variance
$\mathbb{E}[\hat\sigma_r^2 \mid 0 < k < G]$, both of which are maximised
at $p_i = 1/2$ and decay symmetrically. At the configuration used in our
ARPO runs ($G = 8$), the combined factor at $p_i = 0.05$ is approximately
a sixth of its value at $p_i = 0.5$. Tasks the current policy almost
never solves and tasks it almost always solves are systematically
attenuated regardless of $G$ or replay strategy, since replay over
successful trajectories cannot manufacture successes for tasks that have
none.

\textsc{ENVS} computes per-task weights from the full collected dataset
rather than from a current rollout group. Following the loss in \S4.4,
the step weight is
\[
    w_i
    \;=\;
    \mathrm{clip}\!\left(
        \frac{(1 - \mathrm{SR}_i)^{\beta}}{T_i},\; w_{\max}
    \right),
\]
and summing over the $T_i$ step samples of task $i$ gives the per-task
aggregate gradient weight
\[
    \sum_{(s, a) \in \mathcal{D}_i} w_i
    \;=\; T_i \cdot w_i
    \;=\; (1 - \mathrm{SR}_i)^{\beta}
    \quad (\text{pre-clip}),
\]
which is monotonically increasing in task difficulty $1 - \mathrm{SR}_i$
and independent of trajectory length, with the exponent $\beta$
controlling the strength of the correction. Allocation across tasks is
specified in closed form before any gradient step is taken, and the
binomial degeneracy of GRPO does not arise. The trained-set greedy
regression of ARPO-noisy, $54/86$ against $57/86$ for the base policy in
Table~\ref{tab:main-results}, is the behavioural signature of unbalanced
allocation that this construction is designed to avoid.

\subsection{Entropy Collapse}
\label{app:rl:entropy}

For a softmax policy $\pi_n$ updated by policy gradient with step size
$\eta$, \citet{cui2025entropy} show that
\[
    \Delta H_n
    \;=\; -\eta \cdot \mathbb{E}_{s \sim d^{\pi_n}}\!\left[
        \mathrm{Cov}_{a \sim \pi_n(\cdot \mid s)}\bigl(\log \pi_n(a \mid s),\, A^{\pi_n}(s, a)\bigr)
      \right] + O(\eta^2),
\]
where we use $n$ for the update index to avoid collision with the
trajectory step $t$ in \S3.1. They report empirically that this
covariance is positive throughout training across PPO, GRPO, RLOO, and
Reinforce++, so $\Delta H_n \le 0$ at every step under that regime.
Contractions compound because update $n+1$ samples under the
already-narrower $\pi_n$, and online policy gradient cannot increase
action entropy without an explicit entropy bonus.

The supervised loss has the opposite floor. Per state $s$, the
supervised loss is the cross-entropy of $\pi_\theta(\cdot \mid s)$
relative to the empirical $q(\cdot \mid s)$,
\[
    \mathcal{L}_s(\theta)
    \;=\; -\sum_{a} q(a \mid s)\,\log \pi_\theta(a \mid s).
\]
Adding and subtracting $\sum_a q(a \mid s)\,\log q(a \mid s)$ on the
right gives
\[
    \mathcal{L}_s(\theta)
    \;=\; -\sum_a q(a \mid s)\,\log q(a \mid s)
        + \sum_a q(a \mid s)\,\log \frac{q(a \mid s)}{\pi_\theta(a \mid s)},
\]
in which the first term is the Shannon entropy $H[q(\cdot \mid s)]$ and
the second is the KL divergence from $q$ to $\pi_\theta$.
By Gibbs' inequality the KL term is non-negative, with equality iff
$\pi_\theta = q$ on $\mathrm{supp}(q)$, so
\[
    \mathcal{L}_s(\theta)
    \;=\; H[q(\cdot \mid s)] + \mathrm{KL}\bigl(q(\cdot \mid s)\,\big\|\,\pi_\theta(\cdot \mid s)\bigr)
    \;\ge\; H[q(\cdot \mid s)],
\]
attained at the supervised optimum $\pi^* = q$. The converged per-state
entropy therefore equals $H[q]$, which exceeds $H[\pi_0]$ whenever the
empirical distribution $q$ is broader than $\pi_0$ at state $s$. Because
branching retains the $k=3$ highest-agreement action clusters (one on the parent VM and up to two on child VMs)
drawn from $K = 32$ candidates, $|\mathrm{supp}(q)| \le 3$ at any
branching state; when the surviving branches contribute roughly
comparable mass to $q$ (a property of our search-leaf statistics rather
than a design constraint), $H[q] \approx \log_2 |\mathrm{supp}(q)|$,
which exceeds $H[\pi_0]$ on every state where $\pi_0$ is concentrated
below the corresponding entropy ceiling.

Whether this condition is met in practice across our held-out states is
an empirical question. Table~\ref{tab:rl:entropy} reports per-state
action entropy on $54$ held-out states, and the measured ordering is
\[
    H[\pi_{\textsc{envs}}] \;>\; H[\pi_0] \;>\; H[\pi_{\textsc{arpo}}]
    \quad (1.640 > 1.282 > 0.887 \text{ bits}),
\]
matching both halves of the structural argument: \textsc{ENVS} sits
above the base policy, while ARPO sits below it.

\begin{table}[H]
  \caption{Held-out per-state action entropy ($H_{\text{action}}$, in
  bits) and effective number of action categories
  ($2^{H_{\text{action}}}$) on $54$ states ($12$ tasks $\times\ 5$ early
  steps each, stratified by base-policy success-count tier). For each
  policy, $K = 8$ samples at temperature $T = 1$ are drawn at every
  state, and Shannon entropy is computed over the joint action-type and
  quantised spatial-coordinate fingerprint. Pass@8 reproduced from
  Table~\ref{tab:main-results}.}
  \label{tab:rl:entropy}
  \centering
  \small
  \begin{tabular}{lccc}
    \toprule
    Policy            & $H_{\text{action}}$ & $2^{H_{\text{action}}}$ & pass@8 \\
    \midrule
    UI-TARS-1.5-7B    & $1.282$ & $2.43$ & $22.7$ \\
    ARPO-clean        & $0.887$ & $1.85$ & $26.7$ \\
    ENVS-clean        & $1.640$ & $3.12$ & $30.3$ \\
    ENVS-noisy        & $1.729$ & $3.31$ & $29.0$ \\
    \bottomrule
  \end{tabular}
\end{table}

\subsection{Instability: Non-convexity and Drift}
\label{app:rl:stability}

The reinforcement-learning return
\[
    J(\pi)
    \;=\; \sum_\tau \rho_0(s_0)
        \prod_{t=0}^{H-1} \mathcal{T}(s_{t+1} \mid s_t, a_t)\,\pi(a_t \mid s_t)\,
        R_{\mathrm{env}}(\tau),
\]
contains the per-state policy probabilities $\pi(\cdot \mid s)$ as
factors that appear once per visit to state $s$ along $\tau$, so $J$ is
multilinear of degree at most $H$ in the joint per-state simplex. Even
the bilinear case $f(x, y) = xy$ has indefinite Hessian, so $J$ is
non-convex in general. The policy-gradient theorem therefore guarantees
only stationary-point convergence, and PPO/GRPO additionally apply a
non-differentiable importance-ratio clipping at $[1 - \epsilon, 1 +
\epsilon]$ that prevents direct convex analysis. Equilibrium drift
$\mathrm{KL}(\pi_n \,\|\, \pi_0)$ admits no a~priori bound: it is
controlled in practice only by an explicit KL coefficient in the
objective~\citep{schulman2017ppo} or by periodic resets to a reference
policy~\citep{liu2025prorl}, neither of which is part of the ARPO
configuration used in our experiments. \citet{yu2025dapo} report that
production-scale GRPO requires four interventions (Dynamic Sampling,
Clip-Higher, token-level loss, overlong reward shaping) to remain
stable, each addressing an instability that the underlying objective
does not itself control.

The supervised loss is by contrast convex in logit space, with an
explicit a~priori bound on equilibrium drift from $\pi_0$. Writing
$\pi_\theta(a \mid s) = \mathrm{softmax}(z_\theta(s))_a$, the first
derivative of the log-sum-exp $\mathrm{LSE}(z) = \log \sum_b \exp(z_b)$
is $\partial \mathrm{LSE}/\partial z_i = e^{z_i}/\sum_b e^{z_b} = \pi_i$,
and differentiating once more by the quotient rule yields
\[
    H_{ij}
    \;=\; \frac{\partial^2 \mathrm{LSE}(z)}{\partial z_i\, \partial z_j}
    \;=\; \pi_i\,\delta_{ij} - \pi_i \pi_j.
\]
For any $v \in \mathbb{R}^{|\mathcal{A}|}$,
\[
    v^\top H v
    \;=\; \sum_i \pi_i v_i^2 - \Big(\sum_i \pi_i v_i\Big)^2
    \;=\; \mathbb{E}_\pi[v^2] - (\mathbb{E}_\pi[v])^2
    \;=\; \mathrm{Var}_\pi(v) \;\ge\; 0,
\]
so $H$ is positive semi-definite and the per-state cross-entropy
$-z_a + \mathrm{LSE}(z)$ is convex in $z$ (the linear term contributes
zero to the Hessian). The loss is therefore bounded below by $H[q]$ on
the entire logit space, attained at $\pi = q$.

The KL bound at the supervised optimum follows from the same algebra.
For each $a \in \mathrm{supp}(q)$, $q(a)/\pi_0(a) \le \max_b q(b)\,/\,
\pi_0^{\min}(s)$, where $\pi_0^{\min}(s) = \min_{b \in \mathrm{supp}(q)}
\pi_0(b \mid s)$, so
\[
    \mathrm{KL}\bigl(q(\cdot \mid s)\,\big\|\,\pi_0(\cdot \mid s)\bigr)
    \;=\; \sum_a q(a)\,\log_2 \frac{q(a)}{\pi_0(a)}
    \;\le\; \log_2 \frac{\max_a q(a)}{\pi_0^{\min}(s)} \cdot \sum_a q(a)
    \;=\; \log_2 \frac{\max_a q(a)}{\pi_0^{\min}(s)}.
\]
Any cluster retained by \textsc{ENVS} appeared in at least one of $K =
32$ samples drawn from $\pi_0$, so its empirical base frequency is at
least $1/K$. Treating this empirical floor as a working lower bound on
$\pi_0^{\min}(s)$ (a heuristic, since a cluster with true probability
below $1/K$ can survive the retention rule by chance), the per-state KL
at branching states is bounded by approximately $\log_2 K = 5$ bits;
on non-branching states $q$ concentrates near the modal action of
$\pi_0$ and the per-state KL is small. Online RL admits no analogous
a~priori bound on $\mathrm{KL}(\pi_n \,\|\, \pi_0)$, even at convergence;
the trained-set regression $54/86 < 57/86$ in
Table~\ref{tab:main-results} is the training-time signature of this
absence.

\subsection{Rollout Coupling}
\label{app:rl:cost}

Online policy gradient updates the current policy from rollouts collected
under that same policy. Reusing rollouts after the policy has changed
requires importance reweighting, which is not part of the GRPO recipe
implemented by ARPO in our experiments. Each ablation therefore requires
a fresh on-policy collection at the full per-run cost, since past
trajectories were generated under a now-stale policy.

\textsc{ENVS} trajectories are collected under the frozen base policy
$\pi_0$ together with branching, so the dataset is policy-independent
and not specific to any downstream supervised variant. A single search
collection underlies every supervised configuration in this paper at
the cost of an additional supervised pass per variant. The headline
GPU-hour numbers and the data-volume sweep enabled by this amortisation
appear in Table~\ref{tab:compute}.

\section{Limitations}
\label{app:limitations}
Our experiments focus on \textsc{OSWorld} and one base-model family, UI-TARS-1.5-7B. The quantitative results should therefore be read as claims about the studied \textsc{OSWorld} regime, not as universal claims about all GUI agents. Additional experiments on browser, mobile, and other desktop environments are needed to measure transfer.

ENVS also requires infrastructure that is available in \textsc{OSWorld} but not guaranteed in less controlled settings: parallel VM execution, reliable reset, and a terminal oracle for task success. Its behavior fingerprints are practical approximations rather than semantic equivalence classes; two actions with different fingerprints can still be functionally equivalent, and two actions with similar fingerprints can diverge after execution.

\textsc{OSWorld-Noisy} is synthetic and recoverable by design. Its perturbations approximate common desktop interruptions, but they do not cover the full range of real user behavior, system failures, or adversarial interference. Our results also show that noisy collection does not automatically improve \textsc{OSWorld} task completion; perturbations need filtering and balancing to avoid diluting task-solving supervision.

Finally, our ARPO comparisons use the completed configurations in our study, and we do not claim to exhaust all online RL variants or hyperparameter choices. The experiments are expensive, and we do not yet report repeated training seeds for every method. We therefore emphasize matched protocols, compute accounting, and scoped conclusions rather than statistical dominance across all possible runs.

\end{document}